\documentclass{article}

\usepackage[final]{nips_2018}

\usepackage[utf8]{inputenc} 
\usepackage[T1]{fontenc}    
\usepackage{hyperref}       
\usepackage{url}            
\usepackage{booktabs}       
\usepackage{amsfonts}       
\usepackage{nicefrac}       
\usepackage{microtype}      
\usepackage{graphicx}
\usepackage{subcaption}

\newcommand{\cmmnt}[1]{}

\usepackage{todonotes}

\bibpunct{[}{]}{,}{n}{}{;}

\title{Illuminating Generalization in Deep Reinforcement Learning through Procedural Level Generation}

\author{
 Niels Justesen\\
 IT University of Copenhagen\\
 Copenhagen, Denmark \\
 \texttt{noju@itu.dk} \\
\And
 Ruben Rodriguez Torrado\\
 New York University\\
 Brooklyn, USA \\
 \texttt{rrt264@nyu.edu} \\
\And 
 Philip Bontrager\\
 New York University\\
 Brooklyn, USA\\
 \texttt{philipjb@nyu.edu}\\
\And 
 Ahmed Khalifa\\
 New York University\\
 Brooklyn, USA\\
 \texttt{ahmed.khalifa@nyu.edu}\\
\And 
 Julian Togelius\\
 New York University\\
 Brooklyn, USA\\
 \texttt{julian@togelius.com}\\
\And 
 Sebastian Risi\\
 IT University of Copenhagen\\
 Copenhagen, Denmark \\
 \texttt{sebr@itu.dk} \\
}

\begin{document}

\maketitle

\begin{abstract}
Deep reinforcement learning (RL) has shown impressive results in a variety of domains, learning directly from high-dimensional sensory streams. However, when neural networks are trained in a fixed environment, such as a single level in a video game, they will usually overfit and fail to generalize to new levels. When RL models overfit, even slight modifications to the environment can result in poor agent performance. This paper explores how procedurally generated levels during training can increase generality. We show that for some games procedural level generation enables generalization to new levels within the same distribution. Additionally, it is possible to achieve better performance with less data by manipulating the difficulty of the levels in response to the performance of the agent. The generality of the learned behaviors is also evaluated on a set of human-designed levels. The results suggest that the ability to generalize to human-designed levels highly depends on the design of the level generators. We apply dimensionality reduction and clustering techniques to visualize the generators' distributions of levels and analyze to what degree they can produce levels similar to those designed by a human.
\end{abstract}

\section{Introduction}
Deep reinforcement learning (RL) has shown remarkable results in a variety of different domains, especially when learning policies for video games \cite{justesen2017deep}. However, there is increasing evidence suggesting that agents easily overfit to their particular training environment, resulting in policies that do not generalize well to related problems or even different instances of the same problem. Even small game modifications can often lead to dramatically reduced performance, leading to the suspicion that these networks learn reactions to particular situations rather than general strategies~\cite{kansky2017schema,zhang2018study}.

This paper has four contributions. 
\textbf{First}, we show that deep reinforcement learning overfits to a large degree on 2D arcade games when trained on a fixed set of levels. These results are important because similar setups are particularly popular to use as benchmarks in deep reinforcement learning research (e.g.\ the Arcade Learning Environment \cite{bellemare13arcade}). Our findings suggest that policies trained in such settings merely memorize certain action sequences rather than learning general strategies to solve the game.   
\textbf{Second}, we show that it is possible to overcome such overfitting by introducing Procedural Content Generation (PCG) \cite{shaker2016procedural}, more specifically procedurally generated levels, in the training loop. However, this can lead to overfitting on a higher level, to the distribution of generated levels presented during training. This paper investigates both types of overfitting and the effect of several level generators for multiple games. 
\textbf{Third}, we introduce a particular form of PCG-based reinforcement learning, which we call \emph{Progressive PCG}, where the difficulty of levels/tasks is increased gradually to match the agent's performance. 
While similar techniques of increasing difficulty do exist, they have not been combined with a PCG-based approach in which agents are \emph{evaluated on a completely new level every time a new episode begins.} Our approach applies constructive level generation techniques, rather than pure randomization, and this paper studies the effect of several level generation methods.
\textbf{Fourth}, we analyze distributions of procedurally generated levels using dimensionality reduction and clustering to understand whether they resemble human-designed levels and how this impacts generalization.

It is important to note that the primary goal of this paper is not to achieve strong results on human levels, but rather to gain a deeper understanding of overfitting and generalization in deep RL, which is an important and neglected area in AI research. We believe this paper makes a valuable contribution in this regard, suggesting that a PCG-based approach could be an effective tool to study these questions from a fresh perspective. We also see this study relevant for robotics, where an ongoing challenge is to generalize from simulated environments to real-world scenarios.

\section{Related Work}
\label{related_work}
Within supervised learning, it is generally accepted that accuracy (and other metrics) are reported on a testing set that is separate from the training set. In contrast, in reinforcement learning research it is common to report results on the very same task a model was trained on. However, several recent learning-focused game AI competitions, such as the Visual Doom \cite{Kempka2016ViZDoom} AI Competition, The General Video Game AI Learning Track \cite{liusingle,torrado2018deep} and the OpenAI Retro Contest\footnote{\url{https://contest.openai.com/}} evaluate the submitted controllers on levels that the participants did not have access to. None of them are, however, based on procedurally generated levels. The only game AI competition to prominently feature procedural level generation is the Mario AI Competition which did not have provisions for learning agents~\cite{togelius2013mario}.

Randomization of objects in simulated environments has shown to improve generality for robotic grasping to such a degree that the robotic arm could generalize to realistic settings as well \cite{tobin2017domain}. Low-fidelity texture randomization during training in a simulated environment has allowed for autonomous indoor flight in the real world \cite{sadeghi2016cad2rl}. Random level generation has been applied to video games to enable generalization of reinforcement learning agents \cite{beattie2016deepmind,graves2016hybrid,groshev2017learning,bipedalwalker2016klimov}. Several RL approaches exist that manipulate the reward function instead of the structure of the environment to ease learning and ultimately improve generality, such as Hindsight Experience Replay \cite{andrychowicz2017hindsight} and Rarity of Events \cite{justesen2018automated}.

The idea of training agents on a set of progressively harder tasks is an old one and has been rediscovered several times within the wider machine learning context. Within evolutionary computation, this practice is known as incremental evolution~\cite{gomez1997incremental,togelius2006evolving}. 
For example, it has been shown that while evolving neural networks to drive a simulated car around a particular race track works well, the resulting network has learned only to drive that particular track; but by gradually including more difficult levels in the fitness evaluation, a network can be evolved to drive many tracks well, even hard tracks that could not be learned from scratch~\cite{togelius2006evolving}. 
Essentially the same idea has later been independently invented as curriculum learning~\cite{bengio2009curriculum}. Similar ideas have been formulated within a coevolutionary framework as well~\cite{brant2017minimal}.

Several machine learning algorithms also gradually scale the difficulty of the problem. Automated curriculum learning includes intelligent sampling of training samples to optimize the learning progress \cite{graves2017automated}. Intelligent task selection through asymmetric self-play with two agents can be used for unsupervised pre-training \cite{sukhbaatar2017intrinsic}. The POWERPLAY algorithm continually searches for new tasks and new problem solvers concurrently \cite{schmidhuber2013powerplay} and in Teacher-Student Curriculum Learning \cite{matiisen2017teacher} the teacher tries to select sub-tasks for which the slope of the learning curve of the student is highest. Reverse curriculum generation automatically generates a curriculum of start states, further and further away from the goal, that adapts to the agent's performance \cite{florensa2017reverse}.

A protocol for training reinforcement learning algorithms and evaluate generalization and overfitting, by having large training and test sets, was proposed in \cite{zhang2018study}. Their experiments show that training on thousands of levels in a simple video game enables the agent to generalize to unseen levels. Our (contemporaneous) work here differs by implementing an adaptive difficulty progression along with near endless content generation for several complex video games. 

The IMPALA system~\cite{espeholt2018impala} trains a single network to play multiple games simultaneously. Here, the same games were used for training and testing, and it is in principle possible that the network simply learned individual behaviors for all of these games within the shared model.

\section{General Video Game AI Framework}
\label{gvgai}
We are building on the General Video Game AI framework (GVG-AI) which is a flexible framework designed to facilitate the advance of general AI through video game playing \cite{perez2016general}. There are currently over 160 games in GVG-AI which are specified using the declarative video game description language (VGDL)~\cite{schaul2013video}, originally proposed in~\cite{ebner2013towards}. The game definition specifies objects in the game and interaction rules such as rewards and effects of collisions. A level is defined as an ASCII grid where each character represents an object. This allows for quick development of games and levels making the framework ideal for research purposes \cite{perez2018general}.

The GVGAI framework has been integrated with the OpenAI Gym environment \cite{torrado2018deep} which provides a unified RL interface across several different environments \cite{brockman2016openai} as well as state-of-the-art RL implementations \cite{baselines}. While GVG-AI originally provides a forward model that allows agents to use search algorithms, the GVG-AI Gym only provides the pixels of each frame, the incremental reward, and whether the game is won or lost.

\subsection{Parameterized Level Generator}
\vspace{-0.5em}
\label{generator}

\begin{figure*}[!htb]
\begin{center}
  \includegraphics[width=\textwidth]{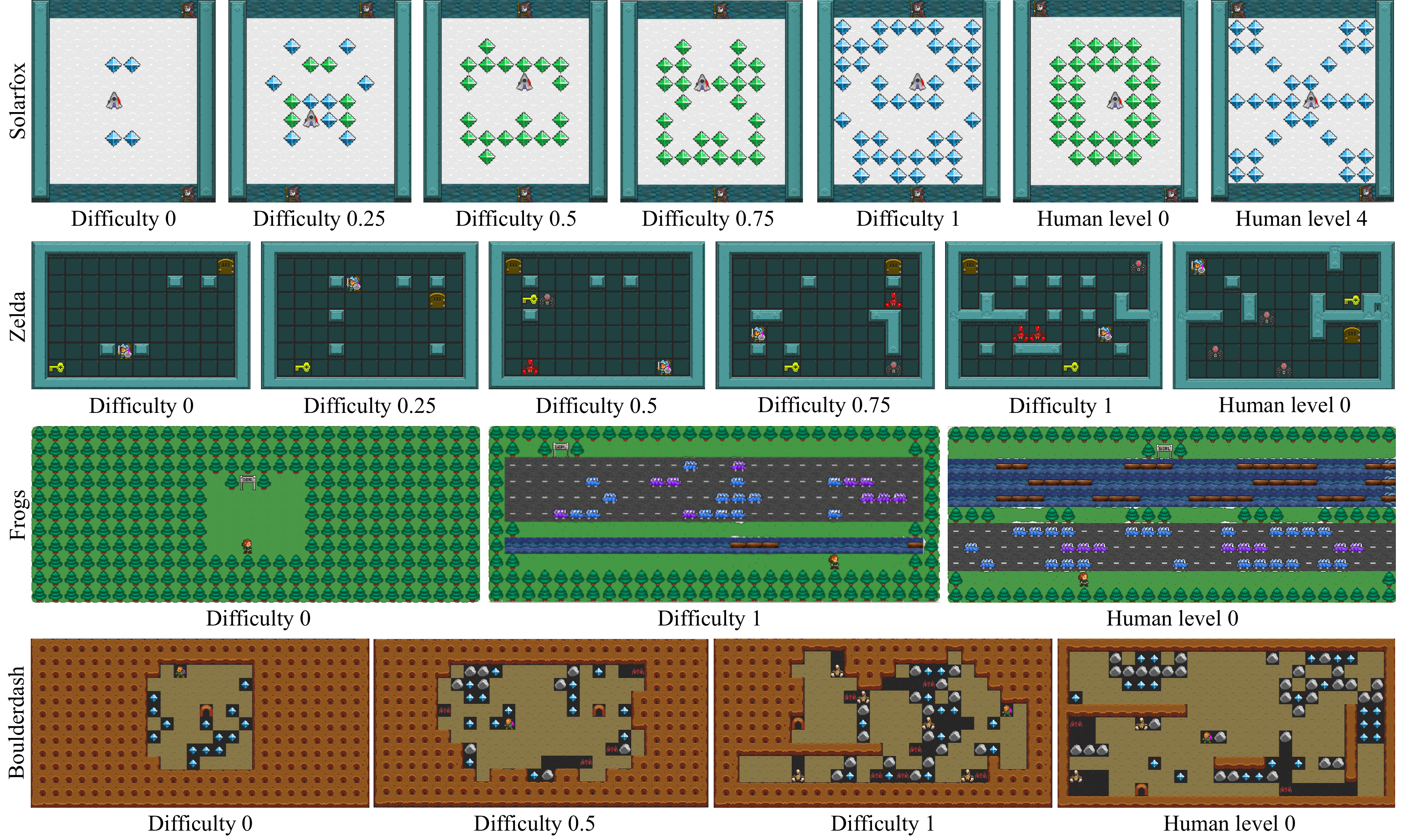} 
  \caption{Procedurally generated levels for Solarfox, Zelda, Frogs, and Boulderdash with various difficulties  between 0 and 1. For each game, human-designed levels are shown as well. } 
  \label{fig:pcg}
  \vspace{-0.5em}
\end{center}
\end{figure*}

Constructive level generators were designed for four hard games in GVG-AI: Boulderdash, Frogs, Solarfox and Zelda. Tree-search algorithms do not perform well in these games~\cite{bontrager2016matching}. 
Constructive level generators are popular in game development because they are relatively fast to develop and easy to debug~\cite{shaker2016procedural}. They  incorporate game knowledge to make sure the output level is directly playable without additional testing. Our generators are designed after analyzing the core components in the human-designed levels for each game and include a controllable difficulty parameter.

\textbf{Boulderdash Level Generator:} This game is a GVG-AI port of ``Boulder Dash'' (First Star Software, 1984). Here the player tries to collect at least ten gems and then exit through the door while avoiding falling boulders and attacking enemies. The level generation in Boulderdash works as follows: (1) Generate the layout of the map using Cellular Automata~\cite{johnson2010cellular}. (2) Add the player to the map at a random location. (3) Add a door at a random location. 
(4) Add at least ten gems to the map at random locations. (5) Add enemies to the map at random locations in a similar manner to the third step.

\textbf{Frogs Level Generator:} Frogs is a GVG-AI port of ``Frogger'' (Konami, 1981). 
In Frogs, the player tries to move upwards towards the goal without drowning in the water or getting run over by cars. The level generation in Frogs follow these steps: (1) Add the player at the lowest empty row in the level. (2) Add the goal at the highest row in the level. (3) Assign the intermediate rows either as roads, water, or forest. (4) Add cars to rows with a road and wood logs rows with water.

\textbf{Solarfox Level Generator:} Solarfox is a GVG-AI port of ``Solar Fox'' (Midway Games, 1981). In Solarfox, the player is continuously moving in one of four directions (North, South, East, and West). The goal is to collect all the gems while avoiding the borders of the level as well as bullets from enemies in the north and the south. 
The level generation for Solarfox follow these steps: (1) Add the player in the middle of the map. (2) Add some gems either in the upper half, left half, or upper left quarter. (3) Replicate the same pattern of gems on the remaining parts of the map.

\textbf{Zelda Level Generator:} Zelda is a GVG-AI port of the dungeon system in ``The Legend of Zelda'' (Nintendo, 1986). In Zelda, the goal is to grab a key and exit through a door without getting killed by enemies. The player can use their sword to kill enemies for higher scores. The level generation in Zelda works as follows: (1) Generate the map layout as a maze using Prim's Algorithm~\cite{buck2015mazes}. (2) Remove some of the solid walls in the maze at random locations. (3) Add the player to a random empty tile. (4) Add the key and exit door at random locations far from the player. (5) Add enemies in the maze at random locations far away from the player.


The difficulty of the levels created by the generator can be controlled with a \emph{difficulty parameter} that is in the interval [0, 1]. Figure~\ref{fig:pcg} shows the effect of the difficulty parameter in the four games. Increasing the difficulty has three effects: First, the area in the level where the player can move through (active level size) increases, except in Zelda and Solarfox where the level size is fixed. Second, the number of objects that can kill the player and/or the number of objects that the player can collect is increased. Third, the layout of the level gets more complex to navigate. The space of possible levels for each game, using our generators, is around $10^8$ at low difficulty to $10^{24}$ at high difficulties. Difficult levels have more possible configurations as they typically have more elements.

\section{Procedural Level Generation for Deep RL}
\label{sec:procedural}

In a supervised learning setting, generality is obtained by training a model on a large dataset, typically with thousands of examples. Similarly, the hypothesis in this paper is that RL algorithms should achieve generality if many variations of the environments are used during training, rather than just one. 
This paper presents a novel RL framework wherein a new level is generated \emph{whenever a new episode begins}, which allows us to algorithmically design the new level to match the agent's current performance. This framework also enables the use of search-based PCG techniques, that e.g.\ learn from existing level distributions \cite{volz2018evolving}, which could in the future reduce the dependency on domain knowledge. However, only constructive PCG is explored in this paper.

When the learning algorithm is presented with new levels continuously during training, it must learn general strategies to improve. Learning a policy this way is more difficult than learning one for just a single level and it may be infeasible if the game rules and/or generated levels have sparse rewards. To ease the learning, this paper introduces \textbf{Progressive PCG} (PPCG),  an  approach where the difficulty of the generated levels is controlled by the  learning algorithm itself. In this way, the level generator will initially create easy levels and progressively increase the difficulty as the agent learns. 


In the PPCG implementation in this paper, levels are initially created with the lowest difficulty of 0. 
If the agent wins an episode, the difficulty will be incremented such that future levels during training become harder. The difficulty is increased by $\alpha$ for a win and decreased by the same amount for a loss. In our experiments, we use $\alpha = 0.01$. For distributed learning algorithms, the difficulty setting is shared across all processes such that the outcome of all episodes influences the difficulty of future training levels. We compare PPCG to a simpler method, also using procedurally generated levels, but with a constant difficulty level. We refer to this approach as \textbf{PCG X}, where \emph{X} refers to the fixed difficulty setting.

\section{Experiments}
\label{experiments}

To evaluate the presented approach, we employ the reinforcement learning algorithm \emph{Advantage Actor-Critic} (A2C) \cite{mnih2016asynchronous}, specifically the implementation of A2C from the Open AI Baselines together with the GVG-AI Gym framework. The neural network has the same architecture as in Mnih et al.~\cite{mnih2016asynchronous} with three convolutional layers and a single fully-connected layer. The output of A2C consists of both a policy and value output in contrast to DQN. A2C is configured to use 12 parallel workers, a step size of $t_{max}=5$, no frame skipping following \cite{torrado2018deep}, and a constant learning rate of $0.007$ with the RMS optimizer \cite{ruder2016overview}. The code for our experiments is available online\footnote{\url{https://github.com/njustesen/a2c_gvgai}}.  


We compare four different training approaches. \textbf{Lv X}: Training level is one of the five human-designed levels. \textbf{Lv 0-3}: Several human-designed levels (level 0, 1, 2, and 3) are sampled randomly during training. \textbf{PCG X}: Procedurally generated training levels with a constant difficulty \emph{X}. \textbf{Progressive PCG (PPCG)}: Procedurally generated training levels where the difficulty is adjusted to fit the performance of the agent. 



\begin{figure*}[!ht]
\centering
\captionsetup[subfigure]{labelformat=empty}
\begin{subfigure}{.245\textwidth}
\includegraphics[width=\textwidth]{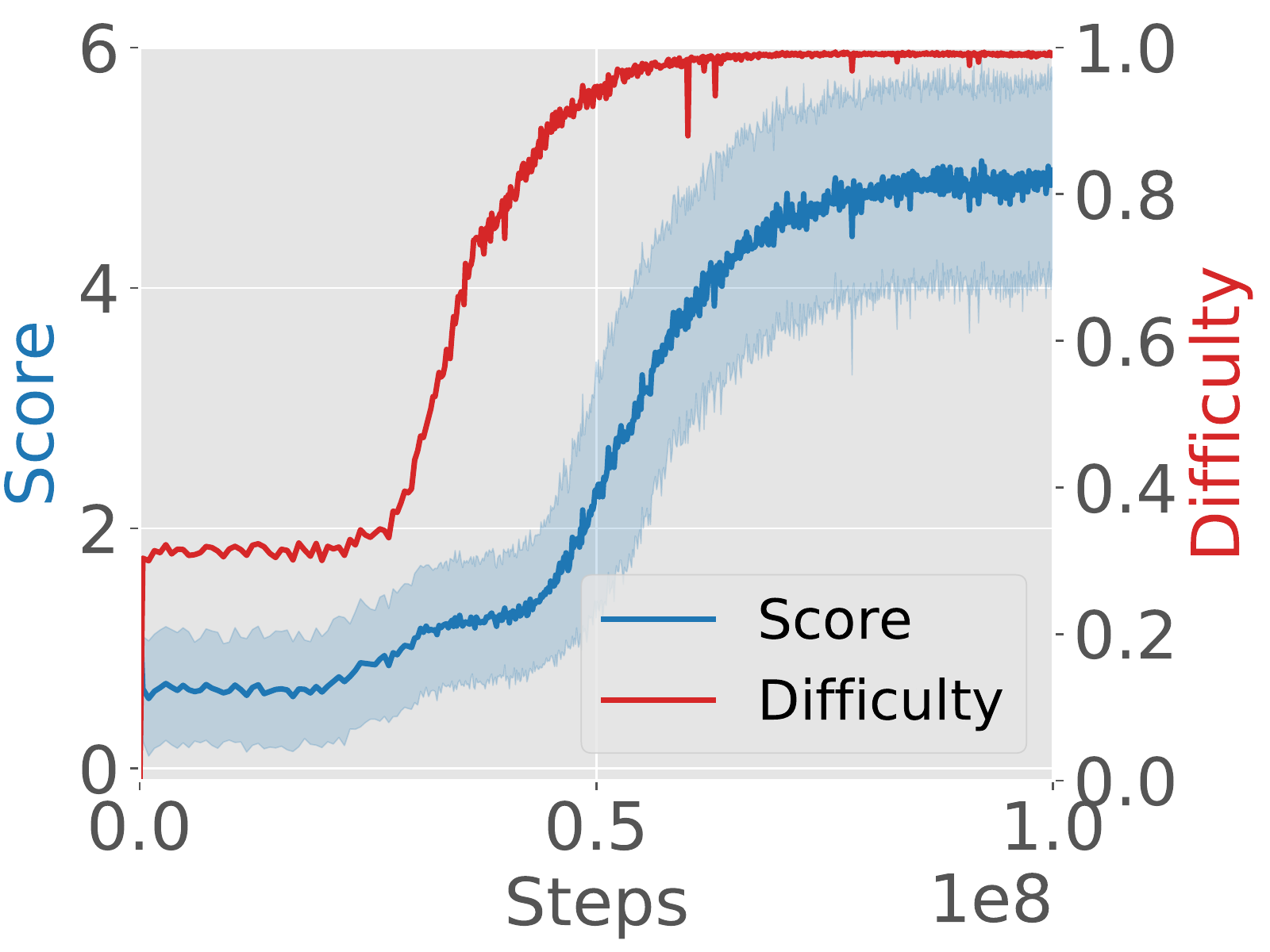}
\caption{(a) PPCG in Zelda}
\end{subfigure}
\begin{subfigure}{.245\textwidth}
\includegraphics[width=\textwidth]{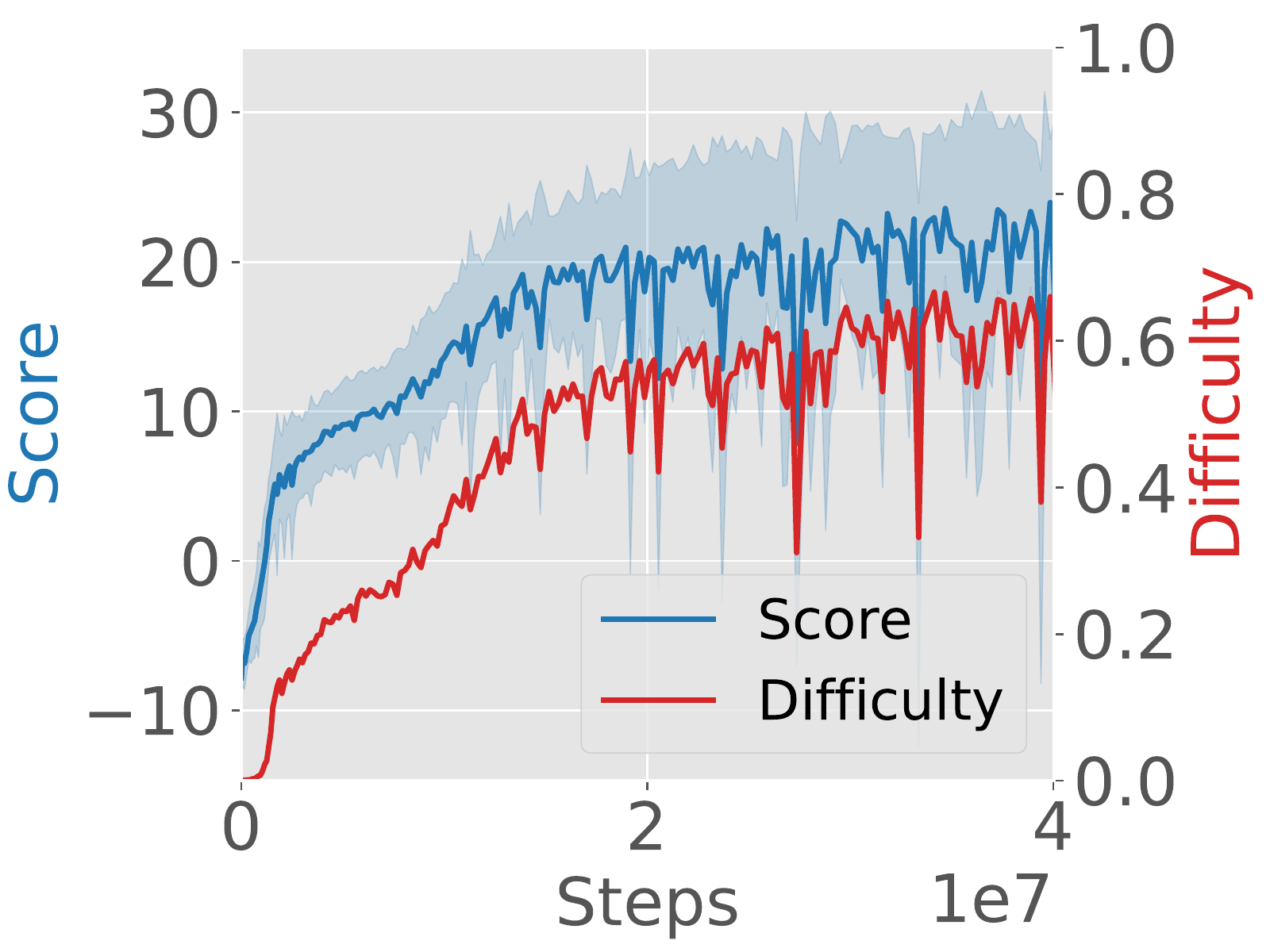}
\caption{(b) PPCG in Solarfox*}
\end{subfigure}
\begin{subfigure}{.245\textwidth}
\includegraphics[width=\textwidth]{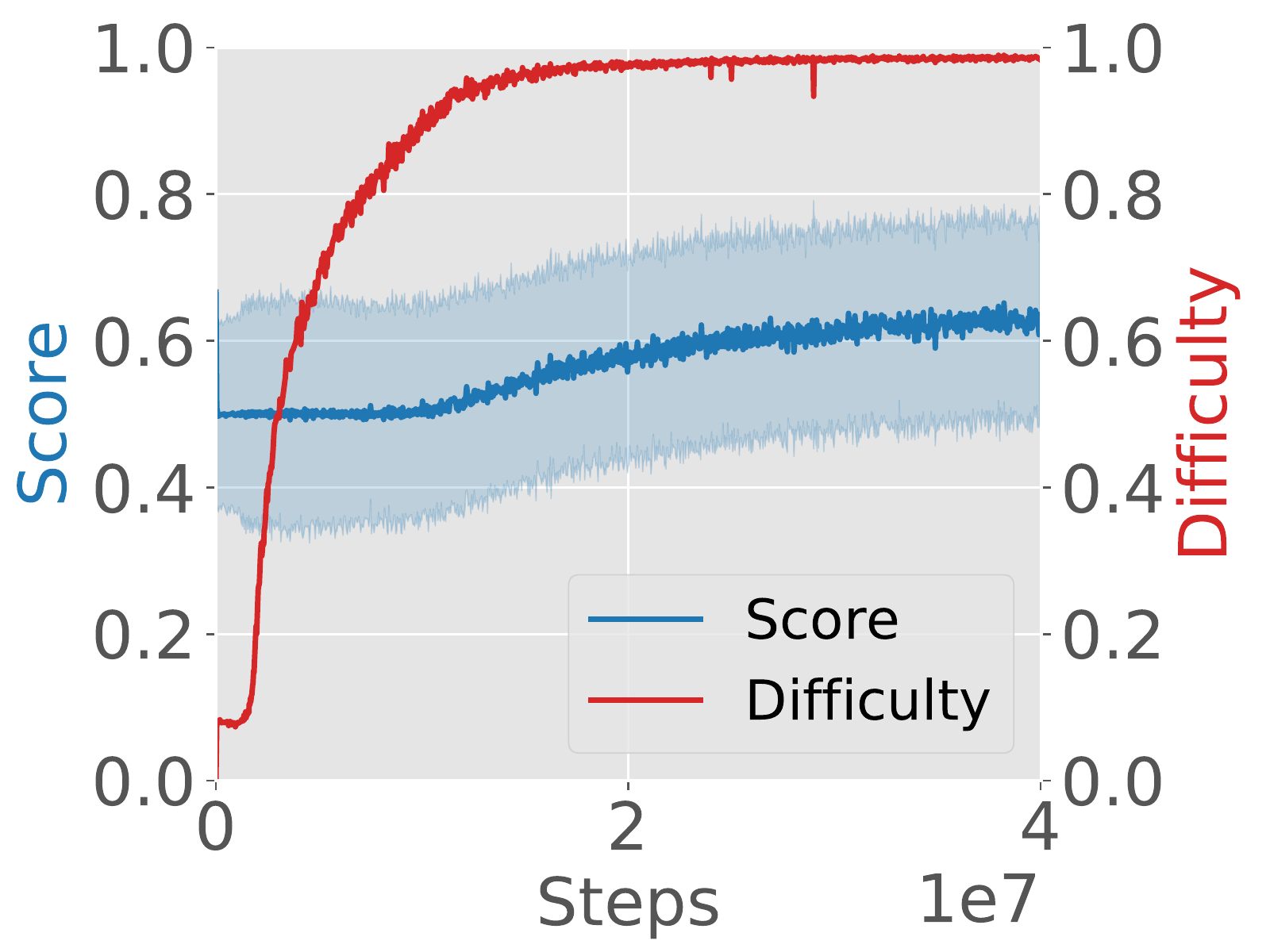}
\caption{(c) PPCG in Frogs}
\end{subfigure}
\begin{subfigure}{.245\textwidth}
\includegraphics[width=\textwidth]{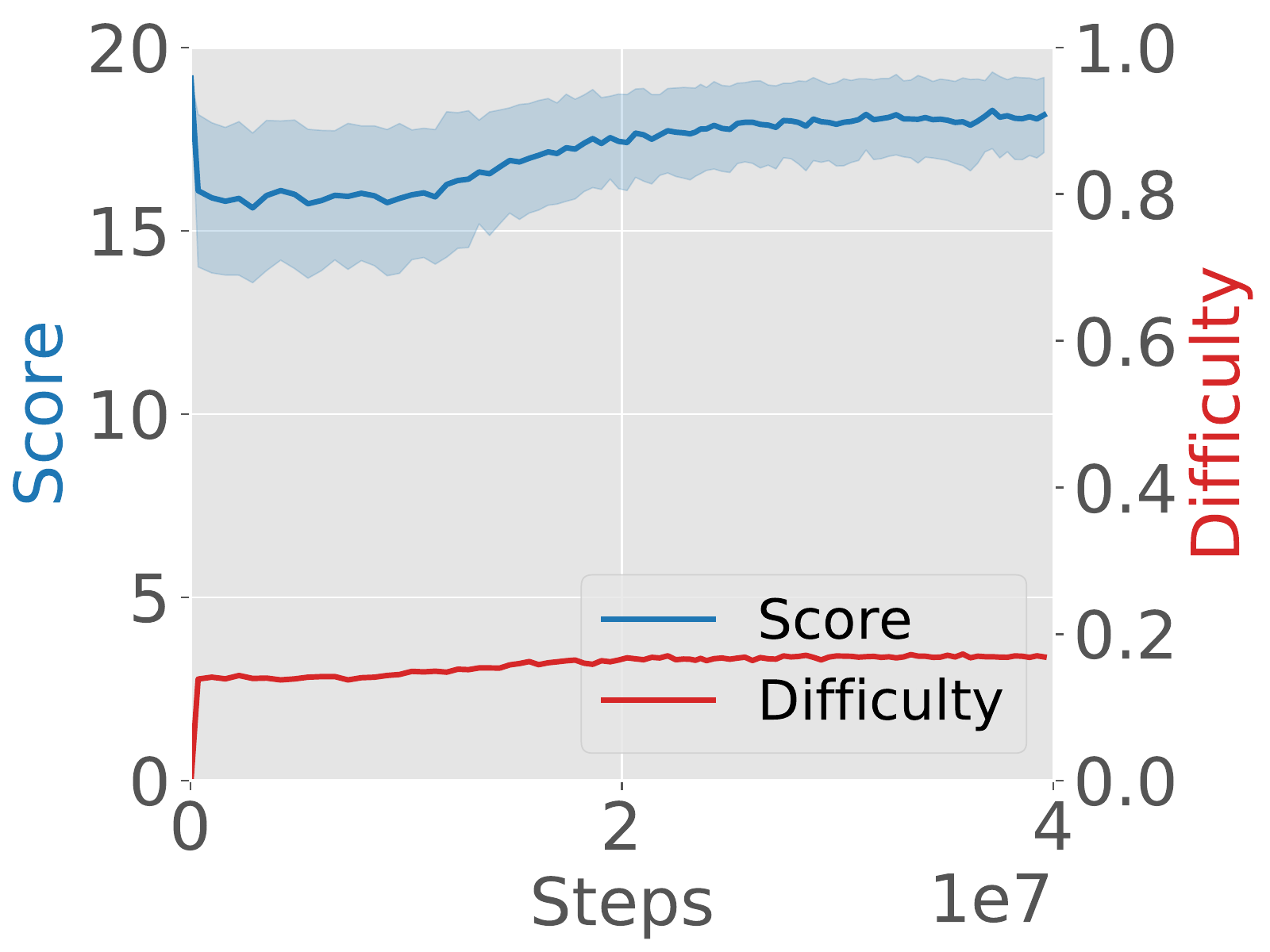}
\caption{(d) PPCG in Boulderdash}
\end{subfigure}%
\caption{Smoothed mean scores and level difficulties during training across five repetitions of Progressive PCG in Zelda, Solarfox, Frogs, and Boulderdash. One standard deviation is shown in opaque. 
*Only three repetitions of PPCG and one of PCG 1 for Solarfox.}
\label{fig:training_plot}
\vspace{-1em}
\end{figure*}

Each training setting was repeated four times and tested on two sets of 30 pre-generated levels with either difficulty 0.5 and 1 as well as the five human-designed levels. The training plots on Figure \ref{fig:training_plot} and the test results in Table~\ref{tab:results} are averaged across the four trained models where each model was tested 30 times on each test setup (thus a total of 120 test episodes per test set for each training setup). All four training approaches were tested on Zelda. Only PCG 1 and PPCG were tested on Solarfox, Frogs, and Boulderdash due to computational constraints. The trained agents are also compared to an agent taking uniformly random actions and the maximum possible score for each test set is shown as well.

\section{Results}

\subsection{Training on a few Human-Designed Levels}
\begin{table}[!htb]
\begingroup
\setlength{\tabcolsep}{6pt} 
\renewcommand{\arraystretch}{1} 
\small
\begin{center}
\begin{tabular}{l|ccccccc }
  \hline
  \multicolumn{8}{c}{Zelda} \\
  \hline
  Training & PCG 0.5 & PCG 1 & Lv 0 & Lv 1 & Lv 2 & Lv 3 & Lv 4 \\
  \hline
  Max.  & 4.40 & 6.87 & 8.00 & 8.00 & 8.00 & 10.00 & 8.00 \\
  Random & 0.38 & 0.22 & 0.26 & 0.17 & -0.11 & -0.07 & 0.18 \\
  \textbf{\scriptsize{60M steps:}}  & & & & & & & \\
  Level 0 & 0.28 & 0.51 & \color{red}6.97 & -0.45 & -0.53 & 0.07 & -0.58 \\
  Level 4 & 0.56 & 0.07 & -0.51 & 0.99 & 0.04 & -0.35 & \color{red}5.93 \\
  Level 0-3  & 1.98 & 2.37 & \color{red}6.95 & \color{red}7.17 & \color{red}7.20 & \color{red}8.17 & \textbf{1.91} \\
  PCG 0.5 & 3.45 & 4.00 & 2.21 & 2.28 & 0.92 & \textbf{2.27} & 0.15 \\
  PCG 1 &0.27 &3.56 &2.40 &1.37 &1.49 &2.88 &-0.62 \\
  PPCG  & 3.44 & 4.28 &2.67 &3.35 &2.43 &1.89 &0.96 \\
  \textbf{\scriptsize{100M steps:}}  & & & & & & & \\
  PCG 1 & 3.05 & 4.38 & 2.49 &1.54 &1.18 & 2.04& -0.29\\
  PPCG  & \textbf{3.82} & \textbf{4.51} & \textbf{2.71} & \textbf{3.74} & \textbf{2.84} & 1.90 & 0.88\\
  \hline
  \multicolumn{8}{c}{Solarfox*} \\
  \hline
  Max.  & 30.83 & 51.83 & 32.00 & 32.00  & 34.00  & 70.00  & 62.00 \\
  Random  & -3.68 & -4.55 & -5.49 & -4.80 ,& -5.41 & 2.03 & 1.13 \\
  \textbf{\scriptsize{40M steps:}}  & & & & & & & \\
  PCG 1 & \textbf{20.70} & \textbf{32.43} & \textbf{22.00} & \textbf{21.83} & \textbf{26.00} & \textbf{43.96} & \textbf{28.16}  \\
  PPCG  &16.08 &21.40 &16.87 &10.26 &12.02 &27.37 & 20.00  \\
  \hline
  \multicolumn{8}{c}{Frogs} \\
  \hline
  Max.  & 1.00 & 1.00 & 1.00 & 1.00 & 1.00 & 1.00  & 1.00 \\
  Random & 0.01 & 0.00 & 0.00 & 0.00 & 0.00 & 0.00 & 0.00 \\
  \textbf{\scriptsize{40M steps:}}  & & & & & & & \\
  PCG 1 & 0.01 & 0.00 & 0.00 & 0.00 & 0.00 & 0.00 & 0.00 \\
  PPCG  & \textbf{0.81} & \textbf{0.57} & 0.00 & 0.00 & 0.00 & 0.00 & 0.00 \\
  \hline
  \multicolumn{8}{c}{Boulderdash} \\
  \hline
  Max.  & 31.50 & 29.80 & 48.00 & 52.00 & 58.00  & 48.00 & 44.00 \\
  Random  & 6.29 & 3.71 & 0.85 & 2.58 & 3.5 & 0.65 & 2.66 \\
  \textbf{\scriptsize{60M steps:}}  & & & & & & & \\
  PCG 1 & \textbf{14.63} & \textbf{8.32} & \textbf{5.39} & \textbf{10.28} & \textbf{5.85} & \textbf{5.08} & \textbf{8.27} \\
  PPCG  & 11.78 & 4.86 & 3.44 & 0.98 & 0.68 & 0.41 & 3.32 \\
  \hline
\end{tabular}
\end{center}
\caption{Test results of A2C under different training regimens: a single human-designed level (\emph{Level 0} and \emph{Level 4}), several human-designed levels (\emph{Level 0-3}), procedurally generated levels with a fixed difficulty (\emph{PCG 0.5} and \emph{PCG 1}), and \emph{PPCG} that progressively adapts the difficulty of the levels to match the agent's performance. \emph{Random} refers to results of an agent taking uniformly random actions and \emph{Max} shows the maximum possible score. 
Scores are in red if the training level is the same as the test level. The best scores for a game, that is not marked red, are in bold. *Only three repetitions of PPCG and one of PCG 1 were made for Solarfox so far.}
\label{tab:results}
\vspace{-2em}
\endgroup
\end{table}

Policies trained on just one level in Zelda (\emph{Lv 0} and \emph{Lv 4} in Table~\ref{tab:results}) reach high scores on the training level but have poor performance on all test levels (human-designed and procedurally generated). It is clear that these are prone to memorization and cannot adapt well to play new levels. The scores on the training levels are close to the maximum scores achievable while the scores on the test levels are often lower than the random policy, a clear indication of overfitting in reinforcement learning. Policies trained on four human-designed levels in Zelda also achieve high scores on all four training levels. The testing scores are marginally higher than when trained on a single level, on both the human-designed level 4 and the PCG levels. 

\subsection{Training on Procedurally Generated Levels}
Agents trained on procedurally generated levels with a fixed difficulty learned a general behavior within the distribution of procedurally generated levels, with mediocre scores in Zelda, Solarfox, and Boulderdash, while no progress was observed in Frogs. These results match similar observations by Torrado et al.~\cite{torrado2018deep}, in which DQN and A2C fail to learn anything on just one level in Frogs after 1 million training steps. While PCG 1, here with 40 million steps, also fails to learn Frogs, PPCG achieves a score of 0.57 (57\% win rate) in the test set of procedurally generated levels with difficulty 1 (comparable to human levels in difficulty - see Figure \ref{fig:pcg}). In Zelda, PCG 1 was able to achieve strong scores while PPCG is slightly better. 
Interestingly, for the two cases where PPCG is able to reach difficulty 1 during training (Frogs and Zelda), it outperforms PCG 0.5 on PCG 1. As PPCG never reaches the most difficult levels during training in Boulderdash and Solarfox, this is to be expected. 
In Boulderdash, the agents trained with PCG 1 reach decent scores (8.34 on average) on levels with difficulty 1. PPCG reached high scores during training but failed to win as the difficulty reached 0.2 and thus trained only on easy levels.

\subsection{Generalization on Human-Designed Levels}
The results demonstrate that introducing procedurally generated levels allows the trained behaviors to generalize to unseen levels within the training distribution. It is, however, interesting whether they also generalize to the five human-designed levels in GVG-AI. 

In Zelda, PCG and PPCG are decent in the human-designed levels while best in the procedurally generated levels.
In Frogs, PCG and PPCG are unable to win in the human-designed levels indicating a clear discrepancy between the two level distributions. 
In Boulderdash, PCG 1 achieved on average 5.08--10.28 points (out of 20) in the human-designed levels compared to 8.32--14.63 in the procedurally generated levels. PPCG performs worse in this game since it never reached a difficulty level similar to the human-designed levels.
Similarly, in Solarfox, PCG 1 achieved on average a higher score than PPCG on the five human-designed levels. PCG 1, however, shows remarkable generalization in Solarfox with similar scores in human-designed and procedurally generated levels.

\subsection{Qualitative Analysis of Agent Replays}
In Zelda, PPCG has learned to reliably strike down and avoid enemies but only sometimes collects the key and exits through the door. Whether this is due to the difficulties of navigating in tricky mazes or a lack of motivation towards the key and door is currently unclear. 
In Solarfox, PCG 1 has learned to effectively pick up the diamonds and avoid fireballs, occasionally getting hit while trying to avoid them. This behavior is remarkably human-like. Sometimes the agent wins in the human-designed levels, which is quite impressive. PPCG jiggles a lot around the starting location to collect nearby diamonds, most likely because the easy procedurally generated levels have diamonds near the starting location, and it never reached the hard levels during training. 
In Frogs, PPCG always moves towards the goal while it sometimes dies when crossing the water with only a few logs being available. We suspect that navigation in this game is learned more easily than in other games as the goal in Frogs is always at the top of the screen. 
In Boulderdash, PCG 1 learned to fight and pick up nearby diamonds, also under boulders, while it does not seem to be capable of long-term planning. It often dies fighting enemies or moving boulders and thus dies rather quickly in most level. Often dying from boulders and enemies can explain why PPCG never reached a difficulty higher than 0.2; it simply gets killed early when these entities are introduced in the levels. 

\section{Exploring the Distribution of Generated Levels}
\begin{figure*}[!htb]
\begin{center}
  \vspace{-.5em}
  \includegraphics[width=\textwidth]{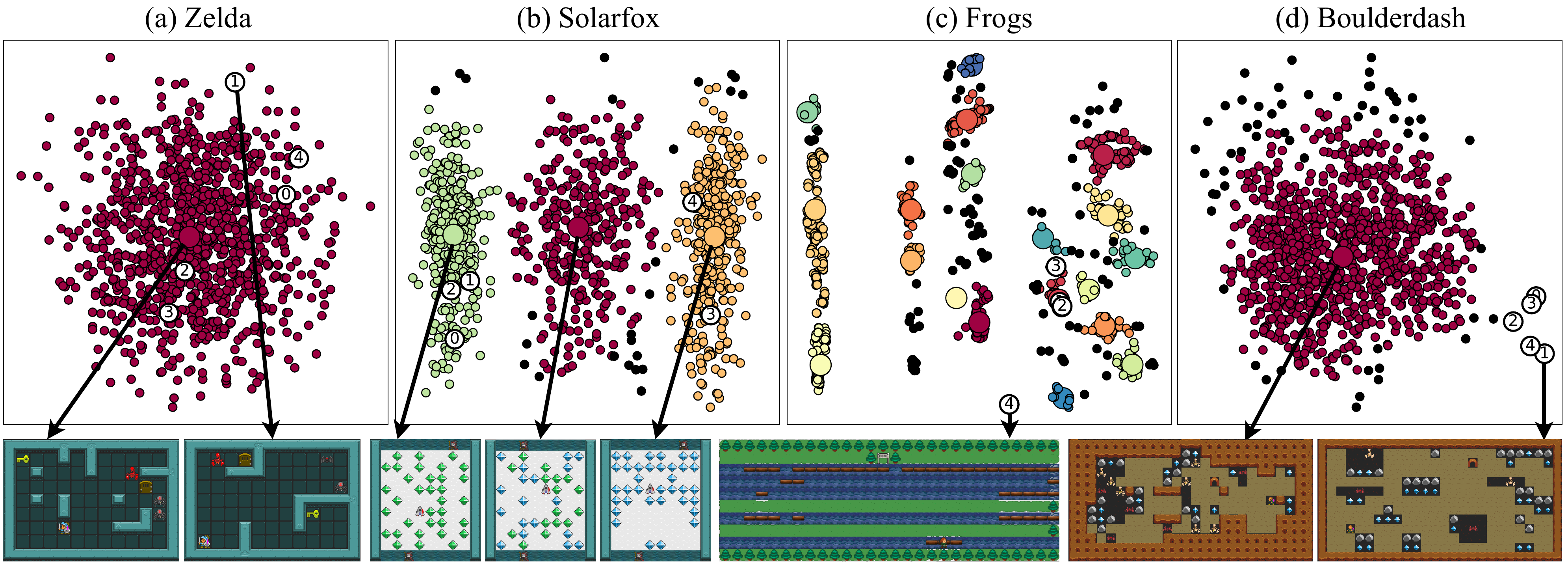} 
  \caption{Visualization of the level distributions and how they correlate to human-designed levels (white circles). Levels were reduced to two dimensions using PCA and clustered using DBSCAN ($\epsilon=0.5$ and a minimum of 10 samples per cluster). Outliers are black and centroids are larger.}
  \label{fig:clustering}
  \vspace{-1em}
\end{center}
\end{figure*}

We do not expect agents to play well on levels that are dissimilar from their training distribution. To investigate the distribution of the procedurally generated levels, and how the structure of these correlate with human-designed levels, we generated 1000 levels with difficulty 1 for each game. The high-dimensional structure of levels was compressed to two dimensions using principal component analysis (PCA), and afterward clustered with the density-based spatial clustering of applications with noise (DBSCAN) approach. The transformed space of levels is visualized in Figure \ref{fig:clustering}. For PCA to work on GVG-AI levels, they have been transformed into a binary 3D array of shape (tile\_type, height, width) and then reshaped into a 1D array. The human-designed levels were included in both the transformation and clustering processes.

The generated levels for Solarfox are clustered in three wide groups: (1) levels with only green diamonds, (2) levels with both green and blue diamonds, and (3) levels with only blue diamonds. None of the human-designed levels use both types of diamonds and thus only belong to two of the clusters. 
For Zelda, only one cluster is discovered without outliers. 
The generated levels in Frogs have been clustered into 19 groups. This is due to the high structural effect of roads and rivers that goes across the level. It is noticeable how level 4 is the most distant outlier. This is because level 4 has a river on the starting row which is a level variation not captured by the level generator for Frogs.  Level 0--3 are near the same small cluster while the generated levels are spread across many isolated clusters. It is not exactly clear why PCG 1 and PPCG fail to play on all the human-designed Frogs levels while the level distribution is remarkably different from the other games. 
In Boulderdash, similarly to Zelda, only one cluster emerges, but here, all human-designed levels are distant outliers. This effect is most likely a result of the fixed amount of open space in the human-designed levels with padding of only one tile while the generated levels are more varied and cave-like. 

\section{Discussion}
\label{discussion}







The results of our experiments affirm the original concern with the way reinforcement learning research is often evaluated and reported. When it is reported that an algorithm has learned a policy that can play a game, it may simply mean that this policy has found optimal actions for a very small subspace of the possible observations the game offers. This boils down to the network mapping observations in this subspace to actions without learning general concepts of the game. Table~\ref{tab:results} shows this with the huge disparity between 
the performance on the training levels compared to the test levels. 
If the goal of the agent is to learn how to play a game, then this work shows that it must be evaluated in several variations of the game. 

Incorporating procedurally generated levels in the training loop also presents a variety of  new and interesting challenges. One such challenge is how to scale the difficulty of the levels to smoothen the learning curve in PPCG. In Frogs, it was very effective to apply padding to easy levels, creating smaller levels in the beginning, while it was not sufficient for Boulderdash. Another challenge is how to ensure that the distribution of procedurally generated levels matches another distribution, in this case human-designed levels. We have provided a tool using dimensionality reduction and clustering, which can be used to improve the design of constructive level generators or perhaps guide search-based level generators in future work. While the results vary across the four games, analyzing when the PCG-based approach works and when it fails gave valuable insights into the generalization abilities of these RL algorithms. We believe that search-based PCG is an interesting area for future work that could ultimately lead to RL agents with more general policies. We believe that this study is also relevant for robotics; learning to generalize from simulation to real-world scenarios where pure randomization of the environment is insufficient. 

\section{Conclusion}
\label{conclusion}

We explored how policies learned with deep reinforcement learning generalize to levels that were not used during training. The results demonstrate that agents trained on just one or a handful of levels often fail to generalize to new levels. This paper presented a new approach that incorporates a procedural level generator into the reinforcement learning framework, in which a new level is generated for each episode. 
%
The presented approach, \emph{Progressive PCG} (PPCG), shows that dynamically adapting the level difficulty during training allows the agent to solve more complex levels than training it on the most difficult levels directly.
This technique was able to achieve a win rate of 57\% in difficult Frogs levels, compared to 0\% for the non-progressive approach. Additionally, in Zelda this approach was superior across procedurally generated levels and human-designed levels. In Solarfox and Boulderdash, the level difficulty of PPCG never reached the maximum during training and here training on procedurally generated levels with a fixed difficulty setting resulted in the highest performance. The results of this paper also highlight the important challenge of ensuring that the training distribution resembles the test distribution. We have provided a tool that can assist with the second challenge, using dimensionality reduction and clustering to visualize the difference between two distributions of video game levels. 

\section*{Acknowledgements}
Niels Justesen was financially supported by the Elite Research travel grant from The Danish Ministry for Higher Education and Science. Ahmed Khalifa acknowledges the financial support from NSF grant (Award number 1717324 - "RI: Small: General Intelligence through Algorithm Invention and Selection.").

\bibliographystyle{abbrv}
\bibliography{references}

\begin{thebibliography}{10}

\bibitem{andrychowicz2017hindsight}
M.~Andrychowicz, F.~Wolski, A.~Ray, J.~Schneider, R.~Fong, P.~Welinder,
  B.~McGrew, J.~Tobin, P.~Abbeel, and W.~Zaremba.
\newblock Hindsight experience replay.
\newblock In {\em Advances in Neural Information Processing Systems}, pages
  5048--5058, 2017.

\bibitem{beattie2016deepmind}
C.~Beattie, J.~Z. Leibo, D.~Teplyashin, T.~Ward, M.~Wainwright, H.~K{\"u}ttler,
  A.~Lefrancq, S.~Green, V.~Vald{\'e}s, A.~Sadik, et~al.
\newblock Deepmind lab.
\newblock {\em arXiv preprint arXiv:1612.03801}, 2016.

\bibitem{bellemare13arcade}
M.~G. {Bellemare}, Y.~{Naddaf}, J.~{Veness}, and M.~{Bowling}.
\newblock The arcade learning environment: An evaluation platform for general
  agents.
\newblock {\em Journal of Artificial Intelligence Research}, 47:253--279, jun
  2013.

\bibitem{bengio2009curriculum}
Y.~Bengio, J.~Louradour, R.~Collobert, and J.~Weston.
\newblock Curriculum learning.
\newblock In {\em Proceedings of the 26th annual international conference on
  machine learning}, pages 41--48. ACM, 2009.

\bibitem{bontrager2016matching}
P.~Bontrager, A.~Khalifa, A.~Mendes, and J.~Togelius.
\newblock Matching games and algorithms for general video game playing.
\newblock In {\em Twelfth Artificial Intelligence and Interactive Digital
  Entertainment Conference}, pages 122--128, 2016.

\bibitem{brant2017minimal}
J.~C. Brant and K.~O. Stanley.
\newblock Minimal criterion coevolution: a new approach to open-ended search.
\newblock In {\em Proceedings of the Genetic and Evolutionary Computation
  Conference}, pages 67--74. ACM, 2017.

\bibitem{brockman2016openai}
G.~Brockman, V.~Cheung, L.~Pettersson, J.~Schneider, J.~Schulman, J.~Tang, and
  W.~Zaremba.
\newblock Openai gym.
\newblock {\em arXiv preprint arXiv:1606.01540}, 2016.

\bibitem{buck2015mazes}
J.~Buck.
\newblock {\em Mazes for Programmers: Code Your Own Twisty Little Passages}.
\newblock Pragmatic Bookshelf, 2015.

\bibitem{baselines}
P.~Dhariwal, C.~Hesse, O.~Klimov, A.~Nichol, M.~Plappert, A.~Radford,
  J.~Schulman, S.~Sidor, and Y.~Wu.
\newblock Openai baselines.
\newblock \url{https://github.com/openai/baselines}, 2017.

\bibitem{ebner2013towards}
M.~Ebner, J.~Levine, S.~M. Lucas, T.~Schaul, T.~Thompson, and J.~Togelius.
\newblock Towards a video game description language.
\newblock 2013.

\bibitem{espeholt2018impala}
L.~Espeholt, H.~Soyer, R.~Munos, K.~Simonyan, V.~Mnih, T.~Ward, Y.~Doron,
  V.~Firoiu, T.~Harley, I.~Dunning, et~al.
\newblock Impala: Scalable distributed deep-rl with importance weighted
  actor-learner architectures.
\newblock {\em arXiv preprint arXiv:1802.01561}, 2018.

\bibitem{florensa2017reverse}
C.~Florensa, D.~Held, M.~Wulfmeier, M.~Zhang, and P.~Abbeel.
\newblock Reverse curriculum generation for reinforcement learning.
\newblock {\em arXiv preprint arXiv:1707.05300}, 2017.

\bibitem{gomez1997incremental}
F.~Gomez and R.~Miikkulainen.
\newblock Incremental evolution of complex general behavior.
\newblock {\em Adaptive Behavior}, 5(3-4):317--342, 1997.

\bibitem{graves2017automated}
A.~Graves, M.~G. Bellemare, J.~Menick, R.~Munos, and K.~Kavukcuoglu.
\newblock Automated curriculum learning for neural networks.
\newblock {\em arXiv preprint arXiv:1704.03003}, 2017.

\bibitem{graves2016hybrid}
A.~Graves, G.~Wayne, M.~Reynolds, T.~Harley, I.~Danihelka,
  A.~Grabska-Barwi{\'n}ska, S.~G. Colmenarejo, E.~Grefenstette, T.~Ramalho,
  J.~Agapiou, et~al.
\newblock Hybrid computing using a neural network with dynamic external memory.
\newblock {\em Nature}, 538(7626):471, 2016.

\bibitem{groshev2017learning}
E.~Groshev, M.~Goldstein, A.~Tamar, S.~Srivastava, and P.~Abbeel.
\newblock Learning generalized reactive policies using deep neural networks.
\newblock {\em arXiv preprint arXiv:1708.07280}, 2017.

\bibitem{johnson2010cellular}
L.~Johnson, G.~N. Yannakakis, and J.~Togelius.
\newblock Cellular automata for real-time generation of infinite cave levels.
\newblock In {\em Proceedings of the 2010 Workshop on Procedural Content
  Generation in Games}, page~10. ACM, 2010.

\bibitem{justesen2017deep}
N.~Justesen, P.~Bontrager, J.~Togelius, and S.~Risi.
\newblock Deep learning for video game playing.
\newblock {\em arXiv preprint arXiv:1708.07902}, 2017.

\bibitem{justesen2018automated}
N.~Justesen and S.~Risi.
\newblock Automated curriculum learning by rewarding temporally rare events.
\newblock In {\em IEEE Conference on Computational Intelligence and Games}.
  IEEE, 2018.

\bibitem{kansky2017schema}
K.~Kansky, T.~Silver, D.~A. M{\'e}ly, M.~Eldawy, M.~L{\'a}zaro-Gredilla,
  X.~Lou, N.~Dorfman, S.~Sidor, S.~Phoenix, and D.~George.
\newblock Schema networks: Zero-shot transfer with a generative causal model of
  intuitive physics.
\newblock {\em arXiv preprint arXiv:1706.04317}, 2017.

\bibitem{Kempka2016ViZDoom}
M.~Kempka, M.~Wydmuch, G.~Runc, J.~Toczek, and W.~Ja\'skowski.
\newblock {ViZDoom}: A {D}oom-based {AI} research platform for visual
  reinforcement learning.
\newblock In {\em IEEE Conference on Computational Intelligence and Games},
  pages 341--348, Santorini, Greece, Sep 2016. IEEE.
\newblock The best paper award.

\bibitem{bipedalwalker2016klimov}
O.~Klimov.
\newblock Bipedalwalkerhardcore-v2.
\newblock \url{http://gym.openai.com/}, 2016.

\bibitem{liusingle}
J.~Liu, D.~Perez-Lebana, and S.~M. Lucas.
\newblock The single-player {GVGAI} learning framework technical manual.
\newblock In {\em IEEE Conference on Computational Intelligence and Games}.
  IEEE, 2018.

\bibitem{matiisen2017teacher}
T.~Matiisen, A.~Oliver, T.~Cohen, and J.~Schulman.
\newblock Teacher-student curriculum learning.
\newblock {\em arXiv preprint arXiv:1707.00183}, 2017.

\bibitem{mnih2016asynchronous}
V.~Mnih, A.~P. Badia, M.~Mirza, A.~Graves, T.~Lillicrap, T.~Harley, D.~Silver,
  and K.~Kavukcuoglu.
\newblock Asynchronous methods for deep reinforcement learning.
\newblock In {\em International Conference on Machine Learning}, pages
  1928--1937, 2016.

\bibitem{perez2018general}
D.~Perez-Liebana, J.~Liu, A.~Khalifa, R.~D. Gaina, J.~Togelius, and S.~M.
  Lucas.
\newblock General video game {AI}: a multi-track framework for evaluating
  agents, games and content generation algorithms.
\newblock {\em arXiv preprint arXiv:1802.10363}, 2018.

\bibitem{perez2016general}
D.~Perez-Liebana, S.~Samothrakis, J.~Togelius, S.~M. Lucas, and T.~Schaul.
\newblock {General Video Game AI: Competition, Challenges and Opportunities}.
\newblock In {\em Thirtieth AAAI Conference on Artificial Intelligence}, pages
  4335--4337, 2016.

\bibitem{torrado2018deep}
R.~Rodriguez~Torrado, P.~Bontrager, J.~Togelius, J.~Liu, and D.~Perez-Liebana.
\newblock Deep reinforcement learning for general video game {AI}.
\newblock In {\em Computational Intelligence and Games (CIG), 2018 IEEE
  Conference on}. IEEE, 2018.

\bibitem{ruder2016overview}
S.~Ruder.
\newblock An overview of gradient descent optimization algorithms.
\newblock {\em arXiv preprint arXiv:1609.04747}, 2016.

\bibitem{sadeghi2016cad2rl}
F.~Sadeghi and S.~Levine.
\newblock Cad2rl: Real single-image flight without a single real image.
\newblock {\em arXiv preprint arXiv:1611.04201}, 2016.

\bibitem{schaul2013video}
T.~Schaul.
\newblock A video game description language for model-based or interactive
  learning.
\newblock In {\em Computational Intelligence in Games (CIG), 2013 IEEE
  Conference on}, pages 1--8. IEEE, 2013.

\bibitem{schmidhuber2013powerplay}
J.~Schmidhuber.
\newblock Powerplay: Training an increasingly general problem solver by
  continually searching for the simplest still unsolvable problem.
\newblock {\em Frontiers in psychology}, 4:313, 2013.

\bibitem{shaker2016procedural}
N.~Shaker, J.~Togelius, and M.~J. Nelson.
\newblock {\em Procedural content generation in games}.
\newblock Springer, 2016.

\bibitem{sukhbaatar2017intrinsic}
S.~Sukhbaatar, Z.~Lin, I.~Kostrikov, G.~Synnaeve, A.~Szlam, and R.~Fergus.
\newblock Intrinsic motivation and automatic curricula via asymmetric
  self-play.
\newblock {\em arXiv preprint arXiv:1703.05407}, 2017.

\bibitem{tobin2017domain}
J.~Tobin, W.~Zaremba, and P.~Abbeel.
\newblock Domain randomization and generative models for robotic grasping.
\newblock {\em arXiv preprint arXiv:1710.06425}, 2017.

\bibitem{togelius2006evolving}
J.~Togelius and S.~M. Lucas.
\newblock Evolving robust and specialized car racing skills.
\newblock In {\em Evolutionary Computation, 2006. CEC 2006. IEEE Congress on},
  pages 1187--1194. IEEE, 2006.

\bibitem{togelius2013mario}
J.~Togelius, N.~Shaker, S.~Karakovskiy, and G.~N. Yannakakis.
\newblock The {Mario AI} championship 2009-2012.
\newblock {\em AI Magazine}, 34(3):89--92, 2013.

\bibitem{volz2018evolving}
V.~Volz, J.~Schrum, J.~Liu, S.~M. Lucas, A.~Smith, and S.~Risi.
\newblock Evolving mario levels in the latent space of a deep convolutional
  generative adversarial network.
\newblock {\em arXiv preprint arXiv:1805.00728}, 2018.

\bibitem{zhang2018study}
C.~Zhang, O.~Vinyals, R.~Munos, and S.~Bengio.
\newblock A study on overfitting in deep reinforcement learning.
\newblock {\em arXiv preprint arXiv:1804.06893}, 2018.

\end{thebibliography}

\end{document}